\documentclass{article}

\usepackage{PRIMEarxiv}

\usepackage[utf8]{inputenc} 
\usepackage[T1]{fontenc}    
\usepackage{hyperref}       
\usepackage{url}            
\usepackage{booktabs}       
\usepackage{amsfonts}       
\usepackage{nicefrac}       
\usepackage{microtype}      
\usepackage{lipsum}
\usepackage{fancyhdr}       
\usepackage{graphicx}       
\graphicspath{{media/}}     
\usepackage{algorithm}
\usepackage{cite}
\usepackage{bm}
\usepackage{amsmath,amssymb,amsfonts}
\usepackage{algorithmic}
\usepackage{graphicx}
\usepackage{subcaption}
\usepackage{multirow}
\usepackage{textcomp}
\usepackage{xcolor}
\title{Trajectory Planning for UAV-Based Smart Farming Using Imitation-Based Triple Deep Q-Learning}
\author{
Wencan~Mao$^{1}$\thanks{The first two authors contributed equally to this work.},
Quanxi~Zhou$^{2}$\footnotemark[1],
Tomás~Couso~Coddou$^{3}$,
Manabu~Tsukada$^{2}$,
Yunling~Liu$^{4}$\thanks{The corresponding author is Yunling Liu (liuyunling@cau.edu.cn).},
Yusheng~Ji$^{1}$\\
$^{1}$National Institute of Informatics, Tokyo, Japan\\
$^{2}$The University of Tokyo, Tokyo, Japan\\
$^{3}$Pontificia Universidad Católica de Chile, Santiago, Chile\\
$^{4}$China Agricultural University, Beijing, China\\
wencan\_mao@nii.ac.jp,
usainzhou@g.ecc.u-tokyo.ac.jp,
tcouso@uc.cl,\\
mtsukada@g.ecc.u-tokyo.ac.jp,
liuyunling@cau.edu.cn,
kei@nii.ac.jp
}

\begin{document}
\maketitle

\begin{abstract}
Unmanned aerial vehicles (UAVs) have emerged as a promising auxiliary platform for smart agriculture, capable of simultaneously performing weed detection, recognition, and data collection from wireless sensors. However, trajectory planning for UAV-based smart agriculture is challenging due to the high uncertainty of the environment, partial observations, and limited battery capacity of UAVs. To address these issues, we formulate the trajectory planning problem as a Markov decision process (MDP) and leverage multi-agent reinforcement learning (MARL) to solve it. Furthermore, we propose a novel imitation-based triple deep Q-network (ITDQN) algorithm, which employs an elite imitation mechanism to reduce exploration costs and utilizes a mediator Q-network over a double deep Q-network (DDQN) to accelerate and stabilize training and improve performance. Experimental results in both simulated and real-world environments demonstrate the effectiveness of our solution.
Moreover, our proposed ITDQN outperforms DDQN by 4.43\% in weed recognition rate and 6.94\% in data collection rate.
\end{abstract}


\section{Introduction}
\label{sec: introduction}

Smart farming represents an innovative approach that integrates information and communication technology into the cyber-physical farm management cycle \cite{big_data_2017}.
Cutting-edge technologies, such as autonomous systems, image processing, machine learning, big data, cloud/edge computing, and wireless sensor networks, have emerged to drive this advancement, paving the way for a better and healthier agricultural practice that features increased production quantity and quality, reduced costs and labor efforts, and lowered fuel, fertilizer, and pesticide utilization \cite{europe_2021}.

While traditional farming relies hugely on human labor, autonomous systems will be able to control actuators effectively, improve the utility, control resource usage, and ensure products conform to market requirements \cite{survey_2021}. 
In particular, unmanned aerial vehicles (UAVs) have been incorporated into smart farming to provide imagery analysis, agricultural surveillance, and in-depth situation awareness.  
Such UAVs are mounted with lightweight cameras, and their applications include insecticide and fertilizer prospecting and spraying, seed planting, weed recognition, fertility assessment, mapping, and crop forecasting \cite{review_2021}.

Under these circumstances, trajectory planning for UAV-based smart farming becomes a critical and challenging issue, especially for large-scale farmlands, where multiple UAVs are supposed to collaborate. First of all, due to the uncertainty in environmental factors,
the distribution and status of the monitoring or managing targets (e.g., seed, crop, weed, and pest) have high spatio-temporal diversity. Without knowledge of such conditions, the effectiveness and efficiency of the smart farming practice will be significantly degraded.
Second, the UAV-mounted cameras have a limited field of view (FoV), namely the maximum area that can be captured by a camera. When the size of the farmland area increases, the UAVs will only have partial observations, meaning that they can only receive incomplete or uncertain information about the environment and the actions of other UAVs. This adds complexity to the decision-making of each individual UAV and their collaboration with each other. 
Besides, the UAVs could have concurrent tasks, such as weed detection, recognition, and data collection from wireless sensors. Such tasks need to be completed simultaneously with their respective requirements.

To address the above challenges, we formulate the trajectory planning problem as a Markov decision process (MDP) and deploy multi-agent reinforcement learning (MARL) to plan the trajectories for UAV-based smart farming.
MARL simulates the interaction among multiple agents and the environment, where each agent is a UAV and the environment refers to the farmland. Unlike traditional optimization methods (e.g., linear programming, A* algorithm, and hyper-heuristics), MARL enables time-efficient policy learning and adaptation to uncertainty in the environment, including the distribution of weeds and wireless sensors, as well as the starting points of the UAVs.

Furthermore, traditional MARL methods, such as Deep Q-Network (DQN), suffer from low scalability and efficiency.
When the size of the field, the amount of weeds, and the number of wireless sensors increase, suboptimal policies can lead to poor task performance or excessive battery consumption.
To address this issue, we propose an \textit{imitation-based triple deep Q-network (ITDQN)}, with a novel design of an \textit{elite imitation mechanism} together with a \textit{mediator Q-network (mid Q-network)} on top of a double deep Q-network (DDQN). The elite imitation mechanism enables the agents to learn from high-performing individuals, thereby reducing exploration costs. Meanwhile, a mediator Q-network is incorporated between the online and target Q-networks, enhancing training performance, efficiency, and stability.

The main contributions of this work are listed as follows:
\begin{itemize}
    \item We target a UAV-based smart farming problem, where UAVs are mounted with lightweight cameras for weed detection and recognition, while collecting information from wireless sensors. We formulate the trajectory planning problem as an MDP to address the high uncertainty in the environment, partial observations of UAVs, and the limited battery of UAVs. 
    \item We propose a novel MARL-based algorithm, referred to as ITDQN, for trajectory planning for UAV-based smart farming.
    Our proposed ITDQN leverages an elite imitation mechanism to lower exploration costs, while incorporating a mediator Q-Network over DDQN to enhance the performance, efficiency, and stability of training.
    \item Extensive evaluations in both simulated and real-world environments validate the effectiveness of our approach. Moreover, the proposed ITDQN consistently outperforms DQN, DDQN, and heuristic baselines.
\end{itemize}

\section{Related Works}
This section reviews the related work from the perspectives of scenarios and algorithms.

\subsection{Trajectory Planning for UAV-Based Smart Farming}

Recent advances in UAV path planning for precision agriculture have explored a diverse range of techniques, each addressing distinct challenges of efficiency, scalability, and adaptability. Energy-efficient flight planning using machine learning has been proposed to minimize energy consumption while maintaining effective coverage~\cite{farm1}. Coverage path planning, which ensures exhaustive scanning of an area of interest, has also been widely studied; however, scalability remains a key limitation, as large agricultural fields may render exhaustive coverage impractical~\cite{farm2,farm2.5}. To reduce computational costs, evolutionary algorithms such as the Grey wolf optimization method have been employed, offering lower-cost solutions compared to traditional optimization approaches~\cite{farm3}. Adaptive path planning methods that integrate deep learning, such as YOLOv4-based object detection combined with Monte Carlo dropout for trajectory design, have further improved responsiveness to environmental uncertainty, though their deterministic policies may limit applicability in multi-UAV settings~\cite{farm4}. Reinforcement learning (RL) has also gained traction, with DQN-based methods enhanced by bidirectional LSTMs demonstrating effectiveness in pest control scenarios~\cite{farm5}. Similarly, Q-learning approaches have been applied to UAV path planning with a focus on shortest-distance priorities, showing advantages over classical algorithms like A* and Dijkstra in single-UAV contexts~\cite{farm6}. Collectively, these studies highlight the growing role of machine learning and optimization in advancing UAV-based agricultural monitoring, while underscoring the need for scalable and multi-agent strategies to fully realize their potential.

\subsection{Q-Learning, DQN, DDQN, and Variants}
As a widely used machine learning method in UAV-based smart farming,
Q-learning \cite{q-learning} is a model-free RL algorithm designed to learn the optimal action-value function through iterative updates of the Bellman equation. Despite its theoretical guarantees, classical Q-learning is limited to discrete and relatively small state-action spaces, as tabular representations become intractable in high-dimensional domains \cite{analysis_dqn}.
Mnih et al. \cite{dqn} addressed the scalability issue by proposing DQN, which employs deep neural networks as approximators for the Q-function. Additionally, DQN introduced two key mechanisms: an experience replay and a target network to stabilize training.
To address the overestimation bias inherent in DQN, Van Hasselt et al. \cite{ddqn} proposed DDQN, which decouples action selection and evaluation by using the online network to select the maximizing action, while the target network evaluates it. This modification reduces overestimation and improves both stability and performance.

Building upon DQN and DDQN, several extensions have been proposed to improve efficiency, exploration, and generalization.
Dueling DQN \cite{duel_dqn} decomposes the Q-function into a state-value component and an advantage component, enabling more robust value estimation in states where action choices have similar outcomes.
Prioritized experience replay (PER) \cite{Schaul2015PrioritizedER} replaces uniform sampling with a priority-based scheme, favoring transitions with larger temporal-difference (TD) errors and thus accelerating learning.
Noisy DQN \cite{noisy_dqn} introduces parameterized noise into network weights, replacing $\epsilon$-greedy exploration with adaptive, learnable stochasticity.
Distributional RL approaches such as C51 \cite{c51}, QR-DQN \cite{qr_dqn}, and IQN \cite{idqn} shift from modeling expected returns to full return distributions, providing richer training signals and empirically stronger performance.
Rainbow DQN \cite{rainbow_dqn} integrates key methods, including DDQN, dueling networks, PER, noisy nets, distributional RL, and n-step returns, into a unified architecture that achieves state-of-the-art results.
Other extensions include ES-DQN \cite{es_dqn}, which improves the bias-variance trade-off of TD targets, and DRQN \cite{dr_dqn}, which incorporates recurrent networks to handle partial observability.

As another extension of DDQN, we propose a novel ITDQN algorithm in this work to improve efficiency, exploration, and stability in trajectory planning for UAV-based smart farming. 

\section{UAV-Based Smart Farming}
This section presents the exemplary scenario, system model, and MDP formulation.
\subsection{Exemplary Scenario}
\begin{figure}[h]
    \centering
\includegraphics[width=0.85\textwidth]{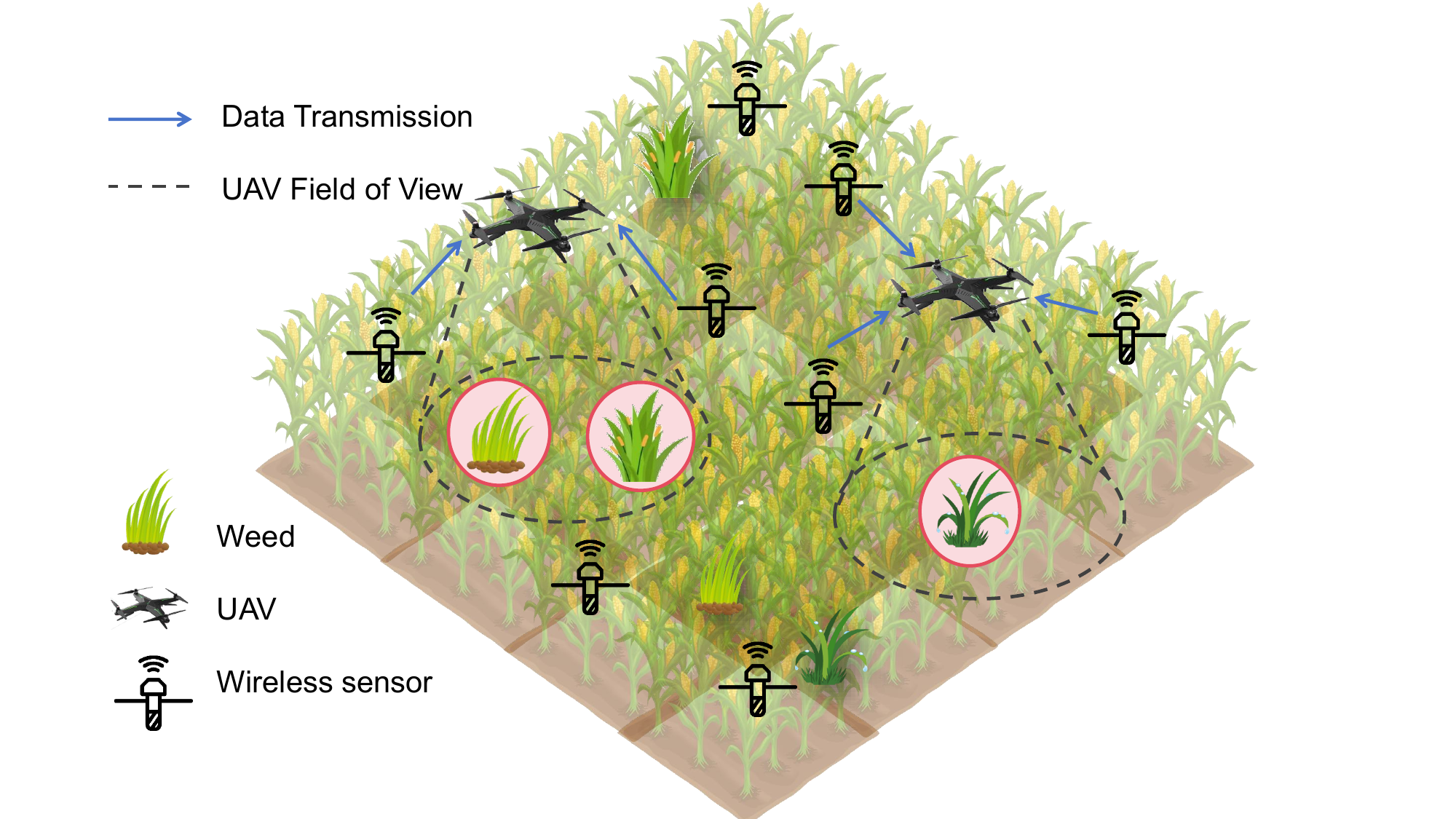}
    \caption{Exemplary scenario of UAV-based smart farming, where UAVs perform weed detection, recognition, and data collection from wireless sensors simultaneously.}
    \label{fig: scenario}
\end{figure}

\begin{figure*}[t]
    \centering
\includegraphics[width=\textwidth,trim={0cm 3.5cm 0cm 7.5cm},clip]{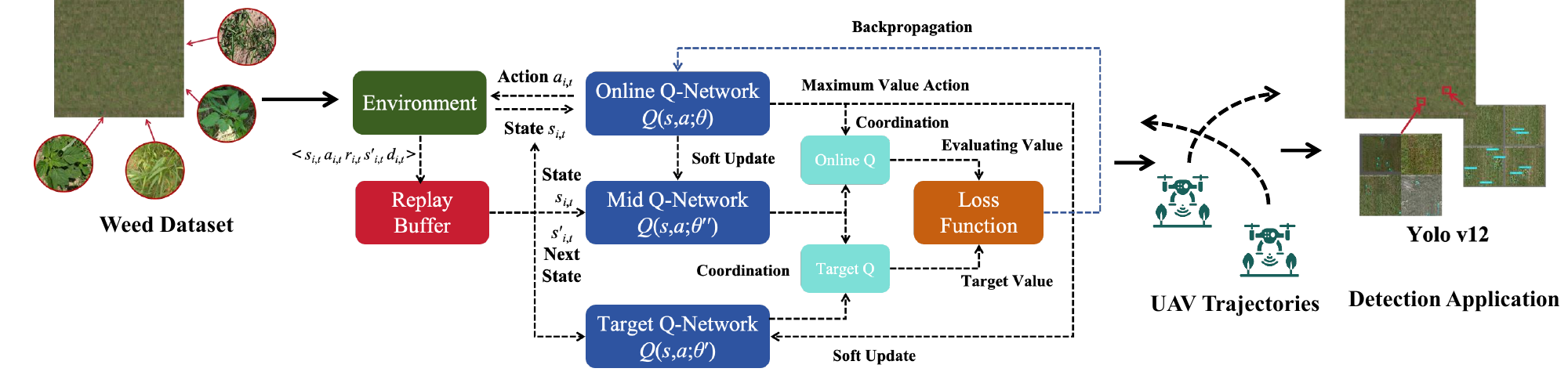}
    \caption{The overview of ITDQN.}
    \label{fig: itdqn.}
\end{figure*}

Due to the side effects and environmental harm of herbicides, reduction of the amounts of herbicides used in farmland is a critical step towards sustainable agriculture \cite{weed_2017}.
In conventional weed control, herbicides are sprayed uniformly across the farmland, treating the soil, crops, and weeds in the same manner. However, this measure will lead to the over-provisioning of herbicides and neglect the difference in herbicide sensitivity for different types of weeds. By mounting lightweight cameras on UAVs, it is possible to detect the distribution of weeds and recognize the type of them, facilitating a smarter and greener weeding process. 

Fig. \ref{fig: scenario} shows an exemplary scenario of smart farming, where $n_{\text{UAV}}$ UAVs are used for weed detection and recognition in a farmland area.
We assume that the UAVs have identical configurations and battery capacity $BC$, and they fly at constant altitude $H$.
For the sake of convenience, we divide the farmland area into $N\times N$ fine-grained grids. 
The total flight duration of the UAVs is an episode $\lambda$, where the UAVs start from random locations. 
During each time step in an episode $t = 1,2,\dots,T_{\text{max}}$, we assume that the UAVs can detect the weeds that are within the FoV. 
It can recognize the density and types of weeds that are within the grid right below it using the embedded object detection algorithm.
Meanwhile, there are $n_{\text{S}}$ randomly distributed wireless sensors mounted on the ground, measuring real-time environmental data (e.g., sunlight, temperature,
humidity, and soil pH).
Apart from weed detection and recognition, the UAVs also collect data from wireless sensors that are within the communication range.
The energy consumption for the UAVs should not exceed their battery capacities in an episode, and they will be fully charged before the start of the next episode.


\subsection{System Model}

\subsubsection{Energy consumption model}

The energy consumption of $\text{UAV}_i$ up to time $t$ can be represented as:
\begin{equation}
E(i,t)=E_{\text{cmp}}(i,t) + E_{\text{cs}}(i,t) + E_{\text{fly}}(i,t),
\end{equation}
where $E_{\text{cmp}}(i,t)$ denotes the cumulative computation energy consumption of $\text{UAV}_i$ at time $t$, $E_{\text{cs}}(i,t)$ denotes the cumulative communication and sensing energy consumption of $\text{UAV}_i$ at time $t$, and $E_{\text{fly}}(i,t)$ denotes the cumulative flight energy consumption of $\text{UAV}_i$ at time $t$.

The computational energy consumption $E_{\text{cmp}}(i,t)$ of UAV $i$ at time $t$ can be represented as:
\begin{equation}
E_{\text{cmp}}(i, t)=\int_{0}^{t} P_{\text{cmp}} \mathrm{d} t,
\end{equation}

\begin{equation}
    P_{\text{cmp}} = P_{\text{static}} + P_{\text{dynamic}},
\end{equation}

\begin{equation}
P_{\text{dynamic}} \approx C^* \cdot V^2 \cdot f^* \cdot \alpha^*,
\end{equation}
where $P_{\text{cmp}}$ represents computational power, $P_{\text{static}}$ represents the static computing module power, $P_{\text{dynamic}}$ represents the dynamic computing module power, $C^*$ represents the load capacitance, $V$ represents voltage, $f^*$ represents clock frequency, and $\alpha^*$ represents the activity factor.

The communication and sensing energy consumption $E_{\text{cs}}(i,t)$ of UAV $i$ at time $t$ can be represented as:
\begin{equation}
E_{\text{cs}}(i, t)=\int_{0}^{t} P_{\text{cs}}\mathrm{d} t,
\end{equation}
where $P_{\text{cs}}$ represents the communication and sensing power.

The flight energy consumption $E_{\text{fly}}(i,t)$ of UAV $i$ at time $t$ can be represented as:

\begin{equation}
E_{\text{fly}}(i, t)=\int_{0}^{t} P_{\text{fly}}(i,t) \mathrm{d} t,
\end{equation}
where $P_{\mathrm{fly}}(i,t)$ denotes the flight power, calculated as:

\begin{equation}
P_{f l y}(i, t)
= c_{1}|\bm{v}(i,t)|^{2}+
c_{2} \frac{{v}_{x}^{2}(i,t)+{v}_{y}^{2}(i,t)}{|\bm{v}(i,t)|^{3}}+m g|\bm{v}(i,t)|
, 
\end{equation}
where $m_{\mathrm{UAV}}$ denotes UAV mass, $g$ is the gravitational acceleration, $\rho_{\mathrm{air}}$ denotes the air density, $A_{\mathrm{UAV}}$ is the UAV frontal area, $\eta$ is the mechanical efficiency, and $\bm{v}(i,t)$ denotes the velocity vector. The components $v_x(i,t)$ and $v_y(i,t)$ represent the UAV's velocity in the $x$ and $y$ directions, respectively. $c_1$ and $c_2$ represent the drag and lift coefficients, respectively, which can be given by:

\begin{equation}
c_{1}=\frac{1}{2} \rho \cdot A_{\text{UAV}} \cdot C_{d},
\end{equation}
\begin{equation}
c_{2}=\frac{m_{\text{UAV}}^{2}}{\eta \cdot \rho_{\text{air}} \cdot n_{\text{prp}} \cdot \pi R_{\text{prp}}^{2}},
\end{equation}
\begin{equation}
A_{\text{UAV}}=A_{\text{surf}}+n_{\text{prp}} \pi R_{\text{prp}}^{2},
\end{equation}
where $C_d$ denotes the air viscosity coefficient, $n_{\mathrm{prp}}$ denotes UAV propeller number, $R_{\mathrm{prp}}$ denotes the radius of the UAV propellers, and $A_{\mathrm{surf}}$ denotes the UAV surface area.

The remaining battery $B(i,t)$ can be presented as:
\begin{equation}
B(i,t) = BC - E(i,t),
\end{equation}
where $BC$ represents the battery capacity.
\subsubsection{Communication model}

Assuming that the packet loss rate of the transmission between $\text{UAV}_i$ and $\text{sensor}_j$ can be represented as:
\begin{equation}
\text{PLR}(i,j) =1-(1-\text{BER}(i,j))^{L},
    \label{18}
\end{equation}
where $L$ represents the packet length, $\text{BER}(i,j)$ represents the bit error rate, with the binary phase shift keying (BPSK) encoding method in this paper, calculated as:
\begin{equation}
\text{BER}\left(i,j\right)=Q\left(\sqrt{2 \cdot \operatorname{SINR}\left(i,j\right)}\right),
    \label{19}
\end{equation}
where $\text{SINR}(i,j)$ represents the signal-to-interference-plus-noise ratio (SINR) between $\text{UAV}_i$ and $\text{sensor}_j$, and $Q(x)$ means Q-function 
of the Gaussian distribution.
They can be represented as:
\begin{equation}
\begin{array}{l}
\operatorname{SINR}(i,t) = \frac{P_{\text{TX}}+G_t-\mathcal{PL}(i,j)}{N_{\text{T}}},
\end{array}
    \label{20}
\end{equation}
\begin{equation}
    \mathcal{PL}(i,j)=20 \log _{10}\left(\frac{4 \pi f_c Dis(i,j)}{c}\right)+0.2 \cdot f_c^{0.3} \cdot Dis(i,j)^{0.6}
\end{equation}

\begin{equation}
\begin{array}{l}
Q(x)=\frac{1}{\sqrt{2 \pi}} \int_{x}^{\infty} e^{-\frac{t^{2}}{2}} d t \approx \frac{1}{2} e^{-\frac{x^{2}}{2}},
\end{array}
    \label{21}
\end{equation}
where $P_{TX}$ represents sensor transmission power,
$\mathcal{PL}(i,j)$ and $Dis(i,j)$ represent the path loss and distance between $\text{UAV}_i$ and $\text{sensor}_j$,
$f_c$ represents carrier frequency, $c$ represents light speed, $n_{\text{S}}$ represents the number of sensors,  
and $N_{\text{T}}$ represents the thermal noise, which can be represented as:
\begin{equation}
N_{\text{T}} = k_B \times T_K  \times Bw,
\end{equation}
where $k_B$ represents Boltzmann constant, $T_K$ represents temperature in Kelvin, and $Bw$ represents bandwidth.

The probability of successful data collection $p_\text{data}(i,j)$ is represented as:
\begin{equation}
p_\text{data}(i,j) = 1 - \text{PLR}(i,j)
\end{equation}
If the data is collected successfully, then the reward $I_{data}$
is received.

\subsection{Markov Decision Process}
As the size of the farmland area scales up, the UAVs with limited FoV will only have a partial observation of the environment. Furthermore, due to the uncertain distribution of weeds and wireless sensors, optimization methods are not feasible for the problem. Therefore, we formulate the trajectory planning problem as the following MDP.
\subsubsection{Agent} Each agent is a UAV that makes its decisions.
\subsubsection{Environment} The environment is the farmland area.
\subsubsection{States $s_{i,t}$} The states of $\text{UAV}_i$ at time $t$ include its location, the existence of weed within the FoV,
the density of the weed, the location and the distance to the nearest wireless sensor that can be connected, the direction towards the highest density of wireless sensors that have been sampled, and the location and the distance to the nearest UAV. At each time $t$, the UAVs will broadcast their locations and the information of the grids they have recognized to other UAVs.
\subsubsection{Actions $a_{i,t}$} The actions of $\text{UAV}_i$ at time $t$ are the flying directions, including North (N), Northeast (NE), East (E), Southeast (SE), South (S), Southwest (SW), West (W), and Northwest (NW).
\subsubsection{Reward $r_{i,t}$} The reward $\text{UAV}_i$ achieve at each time step $t$ is denoted as:
\begin{equation}
    r_{i,t}=P_{out}+P_{bat}+P_{clo}+I_{weed}+I_{data}+I_{exploit}+I_{explore}+b,
\end{equation}
where $P_{out}$, $P_{bat}$, and $P_{clo}$ are penalty terms for flying outside the regional boundary, running out of battery, and being too close to the neighboring UAV, respectively. $I_{weed}$ and $I_{data}$ are rewards for successfully recognizing weeds and collecting data, respectively. $I_{exploit}$ and $I_{explore}$ are incentives for exploitation (flying closer to the nearest sensor) and exploration (flying towards the direction of denser sensors), respectively, and $b$ is a small constant to boost convergence.

\section{Imitation-Based Triple Deep Q Network}

This section proposes a novel MARL-based trajectory planning method, ITDQN, as illustrated in Fig.~\ref{fig: itdqn.}.

\subsection{Imitation-Based Reinforcement Learning}

To enable agents to learn from high-performing individuals and lower the cost for exploration, we propose an \textit{elite imitation mechanism} detailed below. 

We introduce a mimicry cycle factor $\delta$. During episodes determined by $\delta$, each agent $i$ does not update its policy $\pi_i(\theta)$ directly. Instead, it generates an action sequence
$\mathcal{A}_i = \{a_{i,0}, \dots, a_{i,n_{\text{step}}}\}$
and receives the corresponding reward vector $ \mathcal{R}_i = \{r_{i,0}, \dots, r_{i,K_{\text{end}}}\}$, where $n_{\text{step}}$ represents the 
time step in an episode.

The elite (i.e., leading policy) is then identified based on the mean $\mu_i$ and variance $\sigma^2$ of the rewards in $\mathcal{R}_i$. The evaluation metric can be defined as:
\begin{equation}
E\mathcal{R}_i = \beta_1 \mu_i + \beta_2 \sigma_i^2,
\label{19}
\end{equation}
where $\beta_1$ and $\beta_2$ are the weights for mean and variance for elite evaluation, respectively.

Other agents update their policy networks by soft-copying from the leading policy using a soft update parameter for elite $\vartheta$. The update rule is formulated as:

\begin{equation}
\theta_i \gets (1-\vartheta)\theta_i + \vartheta \theta'_{i},
\label{20}
\end{equation}
where $\theta'_{i}$ and $\theta_i$ denote the parameters of the elite and other agents’ policy networks, respectively.

Furthermore, to maintain a certain level of exploration while benefiting from the imitation strategy, we gradually adjust the influence of the imitation process. Specifically, we gradually decrease $\vartheta$ and increase $\delta$.
We initialize the values of $\vartheta$ and $\delta$ as $\vartheta^{*}$ and $\delta^{*}$.
The detailed procedure of this algorithm is shown in Algorithm~\ref{ei}.

\begin{algorithm}[t]
    \caption{Elite Imitation Mechanism.}
    \label{ei}
    \begin{algorithmic}[1]
        \STATE Initialize policy network $\pi_i(\theta)$ for each agent $i$.
        \STATE Initialize $\vartheta\leftarrow\vartheta^{*}$ and $\delta \leftarrow\delta^{*}$.
        \FOR{episode $\lambda = 1, 2, \dots, \lambda_{\text{max}}$}
        \IF{$\lambda \mid \delta$}
        \FOR{agent $i = 1, 2, \dots, n_{\text{UAV}}$}
        \STATE Generate $\mathcal{A}_i$ from $\pi_i(\theta)$. 
        \STATE Obtain Reward $\mathcal{R}_i$ from the interaction between agent $i$ and the environment.
        \ENDFOR
         \STATE Compute the maximum reward evaluation parameter $E\mathcal{R}_i$ as Equation~\eqref{19}.
        \STATE Update $\pi_i(\theta)$ as Equation~\eqref{20}.
        \STATE Update parameters $\vartheta \leftarrow\alpha_1\vartheta$ and $\delta\leftarrow\alpha_2\delta$.
        \ENDIF
        \ENDFOR 
    \end{algorithmic}
\end{algorithm}

\begin{algorithm}[t]
    \caption{Triple Deep Q-Learning.}
    \label{tdqn}
    \begin{algorithmic}[1]
    \STATE Initialize online 
    $Q(s,a;\theta)$ network with weights $\theta$.
    \STATE Initialize target $Q'(s,a;\theta')$ network with weights $\theta' \leftarrow \theta$.
    \STATE Initialize mid $Q''(s,a;\theta'')$ network with weights $\theta'' \leftarrow \theta$. 
    \STATE Initialize replay buffer $\mathcal{D}$.
    \FOR{episode $\lambda = 1,2,\dots,\lambda_{\text{max}}$}
    \FOR{agent $i = 1,2,\dots,n_{{\text{UAV}}}$}
    \STATE Update elite imitation strategy as Algorithm~\ref{ei}.
    \FOR{timestep $t = 1,2,\dots,T_{\text{max}}$}
        \IF{ With probability $\varepsilon$}
        \STATE Select a random action $a$.
        \ELSE
        \STATE Select $a = \arg\max_{a} Q_{\text{online}}(s,a)$ as \eqref{21}.
        \ENDIF
        \STATE Execute action $a_{i,t}$, observe reward $r_{i,t}$, next state $s'_{i,t}$ and task done flag $d_{i,t}$.
        \STATE Store transition $(s_{i,t},a_{i,t},r_{i,t},s'_{i,t},d_{i,t})$ into $\mathcal{D}$.
        \STATE Sample mini-batch $\{(s_j,a_j,r_j,s'_j,d_j)\}$ from $\mathcal{D}$.
        \STATE Compute value target as \eqref{22}.
        \STATE Perform the gradient descent step on loss as \eqref{24}.
        \STATE Update target and mid networks as \eqref{25} and \eqref{26}.
        \ENDFOR
    \ENDFOR
    \ENDFOR
    \end{algorithmic}
\end{algorithm}

\subsection{Triple Deep Q-Learning}

To accelerate learning and improve the accuracy and stability of Q-values, we introduce an additional Q-network, referred to as the \textit{mediator Q-network (mid Q-network)}, over DDQN.

For each agent at each time step, the Q-network and the mid Q-network jointly generate Q-values $Q(s,a;\theta)$ and $Q''(s,a;\theta'')$ for action evaluation, based on which we construct a normal distribution function:
\begin{equation}
    Q_{\text{online}}(s,a) \sim \mathcal{N}(\frac{Q(s,a;\theta)+Q''(s,a;\theta'')}{2}, \sigma^2),
\label{21}
\end{equation}
where $\sigma^2$ indicates a fixed variance.

Subsequently, we sample a tuple $\langle S_{i},A_{i},R_{i},S'_{i},D_i\rangle$ composed of $\langle s_{i,t}$, $a_{i,t}$, $r_{i,t}$, $s_{i,t+1}$, $d_{i,k}\rangle$ from the replay buffer $\mathcal{D}$, where $s_{i,t+1}$ represents the next state and $d_{i,k}$ represents whether task is done at the current time step.

Then the target value $y_{k}$ of the $k$-th sample is expressed as:
    
\begin{equation}
    y_k =  r_{i,k} + (1-d_k)\gamma Q_{\text{target}}(s'_k, \arg\max_{a} Q_{\text{online}}(s'_k,a)),
    \label{22}
\end{equation}
where $Q_{\text{target}}$ represents the evaluation for action value, which can be represented as:
\begin{equation}
    Q_{\text{target}}(s,a) \sim \mathcal{N}(\frac{Q'(s,a;\theta')+Q''(s,a;\theta'')}{2}, \sigma^2).
    \label{23}
\end{equation}

The loss function $L(\theta) $ can be represented as:

\begin{equation}
     L(\theta) = \frac{1}{B} \sum_{k=1}^{B} \big(y_k - Q(s_k,a_k;\theta)\big)^2,
     \label{24}
\end{equation}
where $B$ represents the batch size.

The parameters $\theta'$ and $\theta''$ are soft updated as:
\begin{equation}
    \theta' \leftarrow \tau \theta + (1-\tau)\theta',
    \label{25}
\end{equation}
\begin{equation}
    \theta'' \leftarrow \tau \theta + (1-\tau)\theta'',
    \label{26}
\end{equation}
where $\tau$ represents the soft update parameter for Q-networks.

The detailed procedure of the algorithm is shown in Algorithm~\ref {tdqn}.

\section{Evaluation}
This section presents the experimental setup and results in the simulation and real-world environments, which follow a three-step methodology as depicted in Fig.~\ref{fig: experiment}.

\subsection{Synthetic Simulation}
\subsubsection{Simulation Setup}
\begin{figure}[h]
   \centering
       \begin{subfigure}{0.15\textwidth}
        \centering
        \includegraphics[width=\linewidth]{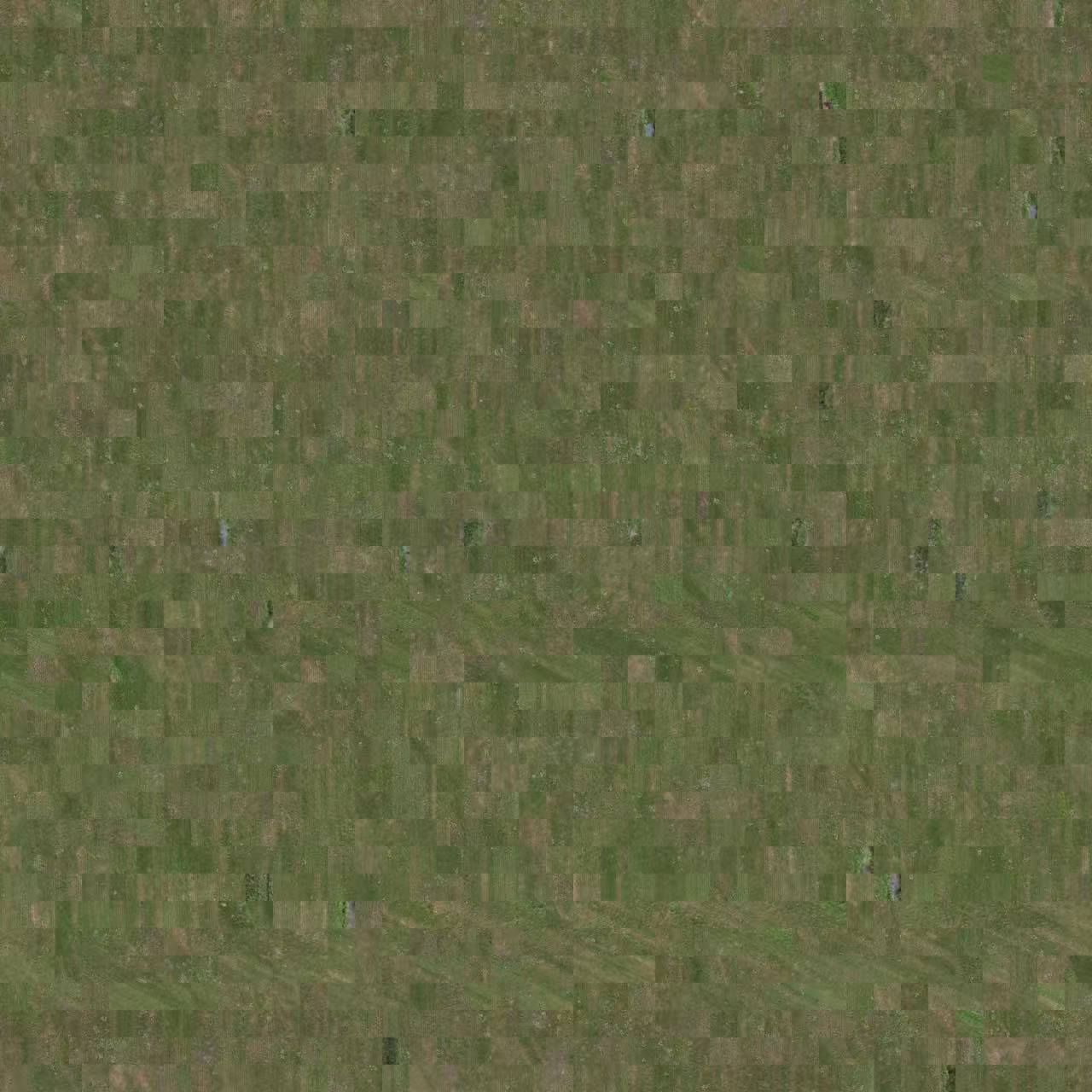}
        \caption{Simulation field.}
        \label{big map}
    \end{subfigure}
    \hfill
    \begin{subfigure}{0.3\textwidth}
        \centering
        \includegraphics[width=\linewidth]{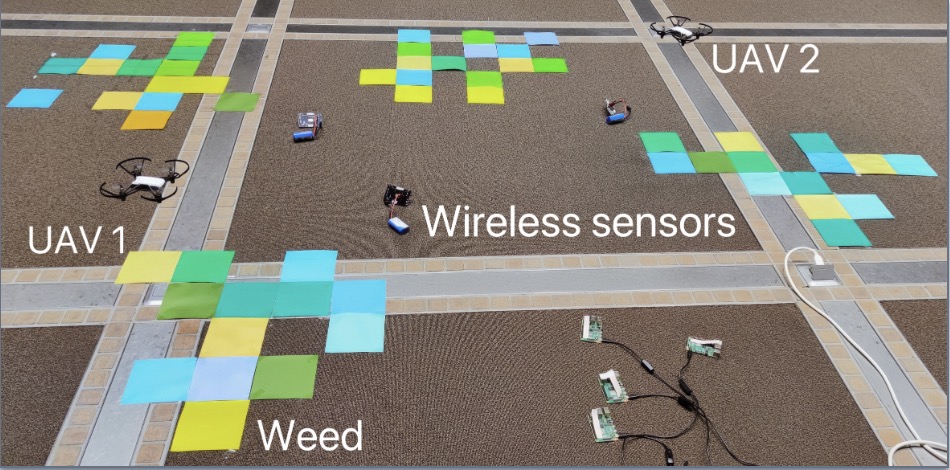}
        \caption{Demo field.}
        \label{test_field}
    \end{subfigure}
     \hfill
    \begin{subfigure}{0.18\textwidth}
        \centering
        \includegraphics[width=\linewidth]{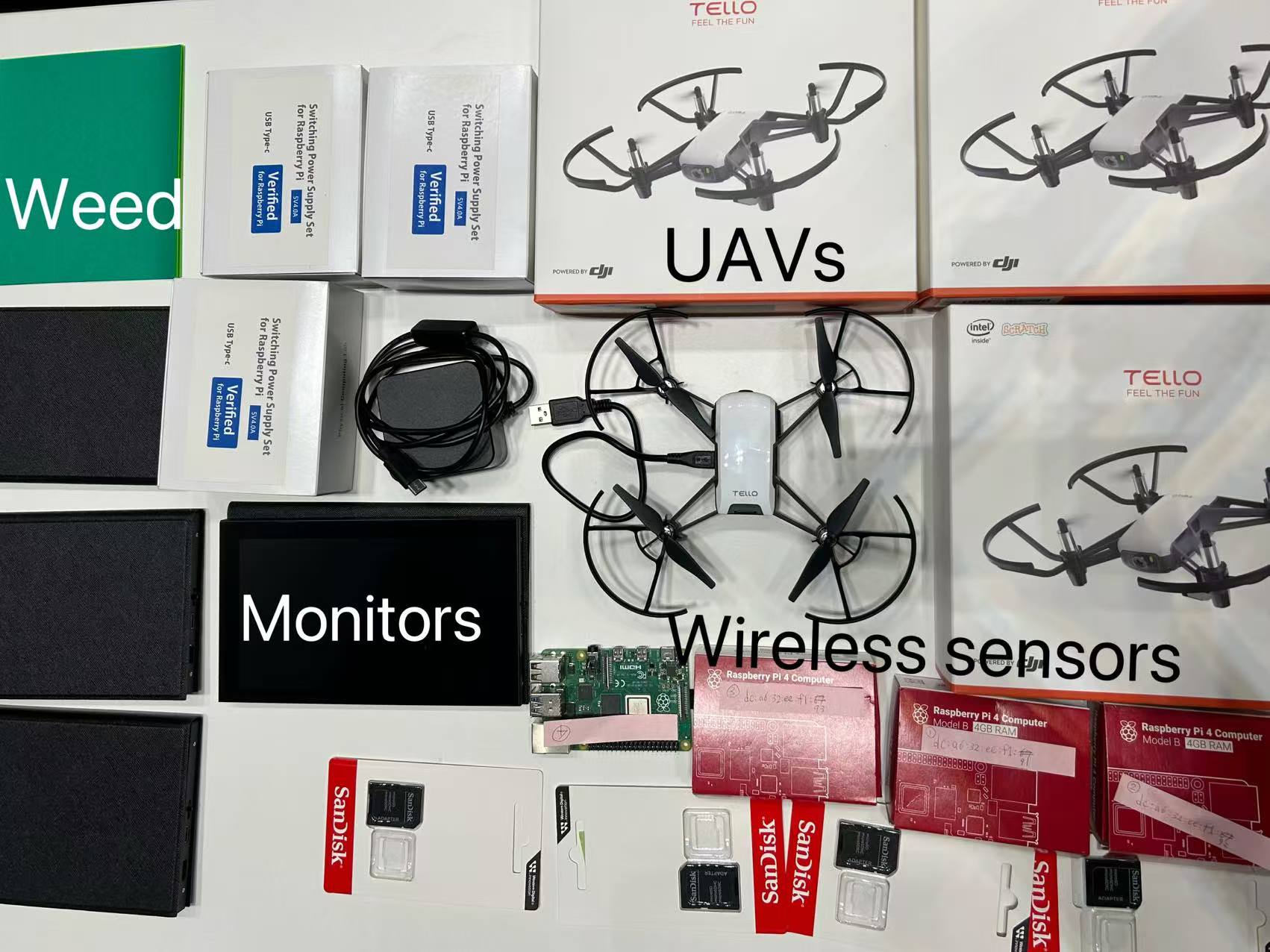}
        \caption{Demo equipment.}
        \label{test_equipment}
    \end{subfigure}
    \hfill
        \begin{subfigure}{0.28\textwidth}
        \centering
        \includegraphics[width=\linewidth]{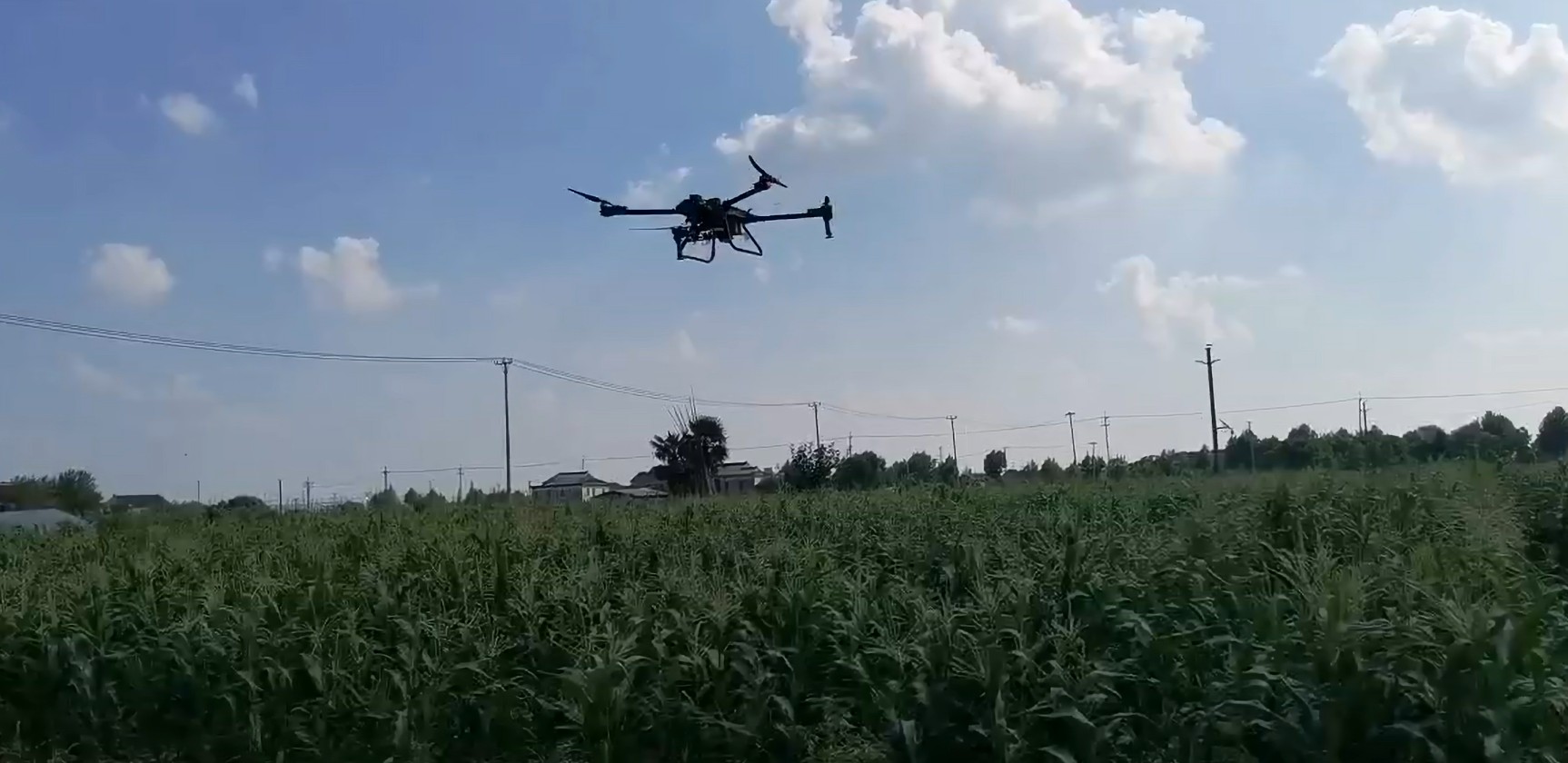}
        \caption{Real-world field.}
        \label{real_field}
    \end{subfigure}
    \hfill
        \caption{Our evaluation follows a three-step methodology as follows. Step~1: Simulation in a synthetic field (a), step~2: Demonstration in an indoor area (b and c), and step~3: Evaluation in a real-world field as our future work (d).}
        \label{fig: experiment}
\end{figure}

In the synthetic simulation shown in Fig.~\ref{big map}, there are 4 UAVs utilized for weed detection and recognition in a farmland area with $20\times 20$ grids, where each grid represents an area of $20~m\times20~m$. Meanwhile, they collect data from 40 wireless sensors randomly generated in the area.
The farmland picture is concatenated using weed images from the Weed Image Detection Dataset \cite{weed_dataset}, 
where each image is annotated with bounding boxes and weed categories for training, and YOLOv12 \cite{yolov12} is used for weed detection and recognition for each UAV.
All measurements are averaged over 10 repeated samples.  Tables~\ref{env_agri} and \ref{hyper_agri} respectively display the environmental and hyperparameters utilized in this work.

\begin{table*}[h]
\caption{Environmental parameters in the experiments.}
\label{env_agri}
\centering
\begin{tabular}{ ||c|c|c|c|c|c|c|c||} \hline
\textbf{Symbol} & \textbf{Definition} & \textbf{Value} &  \textbf{Unit} & \textbf{Symbol} & \textbf{Definition} & \textbf{Value} &  \textbf{Unit} \\
\hline
$N\times N$ & Size of farmland & $20\times 20$ & grids & $n\times n$ & Size of grid & $20\times 20$ & $\text{m}^2$
\\
$FoV$ & UAV field of view & $3\times 3$ & grids & $d_\text{UAV}$ & UAV distance threshold & $20\times \sqrt{2}$ & m \\
$n_\text{UAV}$ & Number of UAVs & $4$&-& $n_{\text{S}}$ & Number of wireless sensors & $40$& -\\
$H$ & UAV flying altitude & 20 & m & $L$ & Packet length & 20 & Byte \\
$BC$ & Battery capacity& $51840$& $\text{J}$ & $a_{max}$ & UAV max acceleration & $20\times\sqrt{2}$ & $\text{m/s}^2$\\
$m_{\text{UAV}}$ & Mass of UAV & 1 & kg & $f_{c}$ & Signal frequency & 2.8 & GHz \\
$g$ & Gravitational acceleration & 9.8&- & $\rho_{\text{air}}$ & Air density & 1.225 & $\text{kg/m}^3$\\
$v_{\text{th}}$& Hovering speed threshold & 0.1 &$\text{m/s}$ & $C_d$ & Viscosity coefficient& 0.5 & $-$\\
$n_{\text{prp}}$ &Propeller number & 4 &- & $R_{\text{prp}}$& Propeller radius& 0.1 & m\\
$\eta$ &Mechanical efficiency & 0.8 & - & $A_{\text{surf}}$& UAV fuselage area & 0.01 & $\text{m}^2$\\
$T_{max}$ & Length of an episode & 200 & s  & $P_\text{static}$ & Static power & 4 & W \\
$t$ & Time step in an episode & 1 & s & $k_B$ &  Boltzmann constant&$1.38 \times 10 ^{-23}$&- \\
$T_K$ &Temperature in Kelvin& 298 & K& $Bw$ & Bandwidth & 20 &MHz \\
$P_{\text{cs}}$ & Commu. sensing power & 30 & dBm & $P_{TX}$ & Sensor transmission power & 17 & dBm \\
{$\alpha^*$} & {Activity factor} & {0.5} & - &
{$C^*$} & {Load capacitance} & {6.4}  & {nF} \\
$c$ & Speed of light & $3\times10^8$ & $m/s$ &
{$f^*$} & {Clock frequency} & {$200 \times 10^6$} & {Hz}\\
{$V$} & {Voltage} & {5} & {V} &&&&\\
\hline
\end{tabular}
\end{table*}

\begin{table*}[h]
\caption{Hyperparameters in the experiments.}
\label{hyper_agri}
\centering
\begin{tabular}{ ||c|c|c|c|c|c||} \hline
\textbf{Symbol} & \textbf{Definition} & \textbf{Value}  & \textbf{Symbol} & \textbf{Definition} & \textbf{Value} \\
\hline
$P_{out}$ & Penalty for flying out of farmland & 10 & $P_{bat}$ & Penalty for battery outage & 10 \\
$P_{clo}$ & Penalty for collision avoidance & $0.1\times \Delta d$ & $b$ & Constant to boost convergence & 0.1 \\
$I_{weed}$ & Reward for weed recognition & 2 & $I_{data}$ & Reward for data collection & 2 \\
$I_{exploit}$ & Incentive for exploitation & 0.05 & $I_{explore}$ & Incentive for exploration & 0.1 \\
$\gamma$ & Discounted factor& 0.99 & $M$ & Memory size & $2^{16}$\\
$r$ & Learning rate & $10^{-4}$ & $D$ & Replay buffer size & $2^{16}$ \\
$B$ & RL training batch size & 128 & $\epsilon$ & Initial exploration probability & 1 \\
$\epsilon_{min}$ & Minimum exploration probability & 0.01 & $\epsilon_{decay}$ & Exploration probability decay rate & 0.995 \\
$\lambda_{\text{max}}$ & Maximum training episode & 1000 &  $hid$ & Hidden dimension & 256\\
$\vartheta^{*}$ & Initial elite soft update parameter & 0.1 &  $\delta^{*}$ & Initial mimicry cycle factor & 10 \\
$\alpha_1$ & Elite soft update parameter decay rate & $1/2$ & $\alpha_2$ & Mimicry cycle factor increase rate & 2 \\
$\beta_1$ & Weight of mean for elite evaluation & 1 & $\beta_2$ & Weight of variance for elite evaluation & 0.01 \\
$\tau$ & Soft update parameter for Q-network & 0.01 & $\sigma^2$ & Fixed variance for Q-network & 0.01 \\
\hline
\end{tabular}
\end{table*}

\subsubsection{Weed Detection and Comparison}
Fig.~\ref{result_yolo} shows the weed detection and recognition results in six exemplary grids with the weed type ``ridderzuring". It can be seen that most of the weeds are successfully detected by the bounding boxes and categorized with correct labels.
The inference time using YOLOv12 is 2.73 milliseconds.
Fig.~\ref{performance_yolo} further details the performance metrics, which yield precision, recall, F1-score, and IoU values above 70\%. 

\subsubsection{Comparison with Baselines}
In the comparative study, we choose three heuristics for comparison, namely ant colony optimization (ACO)~\cite{aco}, particle swarm optimization (PSO)~\cite{pso}, and genetic algorithm (GA)~\cite{ga}.   
As MARL baselines, we choose DQN~\cite{dqn}, and DDQN~\cite{ddqn} for comparison.

Fig.~\ref{comparison} compares the reward versus the number of episodes of DQN, DDQN, and ITDQN. 
Among these algorithms, DQN has the slowest convergence speed and lowest stability. In contrast, DDQN addresses the overestimation bias by duplicating the DQN architecture and redefining the Q-learning update equation.
Thus, it achieves comparable reward with DQN, but with significantly higher speed and less fluctuation.
Moreover, our proposed ITDQN achieves the highest convergence speed because the novel elite imitation mechanism enables the rapid learning and refinement of policies for multiple UAVs. 
Furthermore, the improved triple Q-network architecture facilitates the coordination between online and target Q-networks, surpassing existing algorithms in performance and stability.
Therefore, we can conclude that our proposed ITDQN is superior to the existing DQN and DDQN methods in terms of performance, efficiency, and stability.

Performance metrics obtained by various algorithms are listed in Table~\ref{performance_metrics}. 
Compared with heuristics such as ACO, PSO, and GA, our proposed ITDQN achieves superior performance in data collection and inferior
performance in weed recognition.
The heuristics are used under the assumption of pre-known environments.
However, due to the uncertain distribution of weeds and the random starting points of UAVs, the heuristics are infeasible to deploy in real-world environments.
In addition, compared with MARL algorithms, heuristics have much higher inference times.
The heuristics rely on complex rule-based logic or search procedures, which are computationally intensive. In contrast, MARL policies, once trained, are typically executed through efficient neural network inference, allowing for scalable and real-time decision-making.

Compared to DQN and DDQN, which exhibit similar performance, our proposed ITDQN has higher energy consumption and task completion time, but is still within the battery capacity and episode length requirements. The inference time of DQN is lower than that of DDQN, and both are lower than ITDQN, as they use 1, 2, and 3 layers of Q-networks, respectively. Most importantly, ITDQN achieves the highest task performance, surpassing DDQN by 4.43\% in weed recognition rate and 6.94\% in data collection rate.

\subsubsection{UAV Trajectories}
The visualization of UAV trajectories in the simulation is depicted in Fig.~\ref{UAV_trajectory_simulation}.
The UAV trajectories are aligned with the weed distribution shown in the heatmap. Furthermore, the four UAVs collaborate well for weed detection, recognition, and data collection, with each of them responsible for a cluster of weeds and a data collection region.

\subsection{Real-World Demonstration}

\begin{figure*}[t]
   \centering
       \begin{subfigure}{0.4\textwidth}
        \centering
        \includegraphics[width=\linewidth]{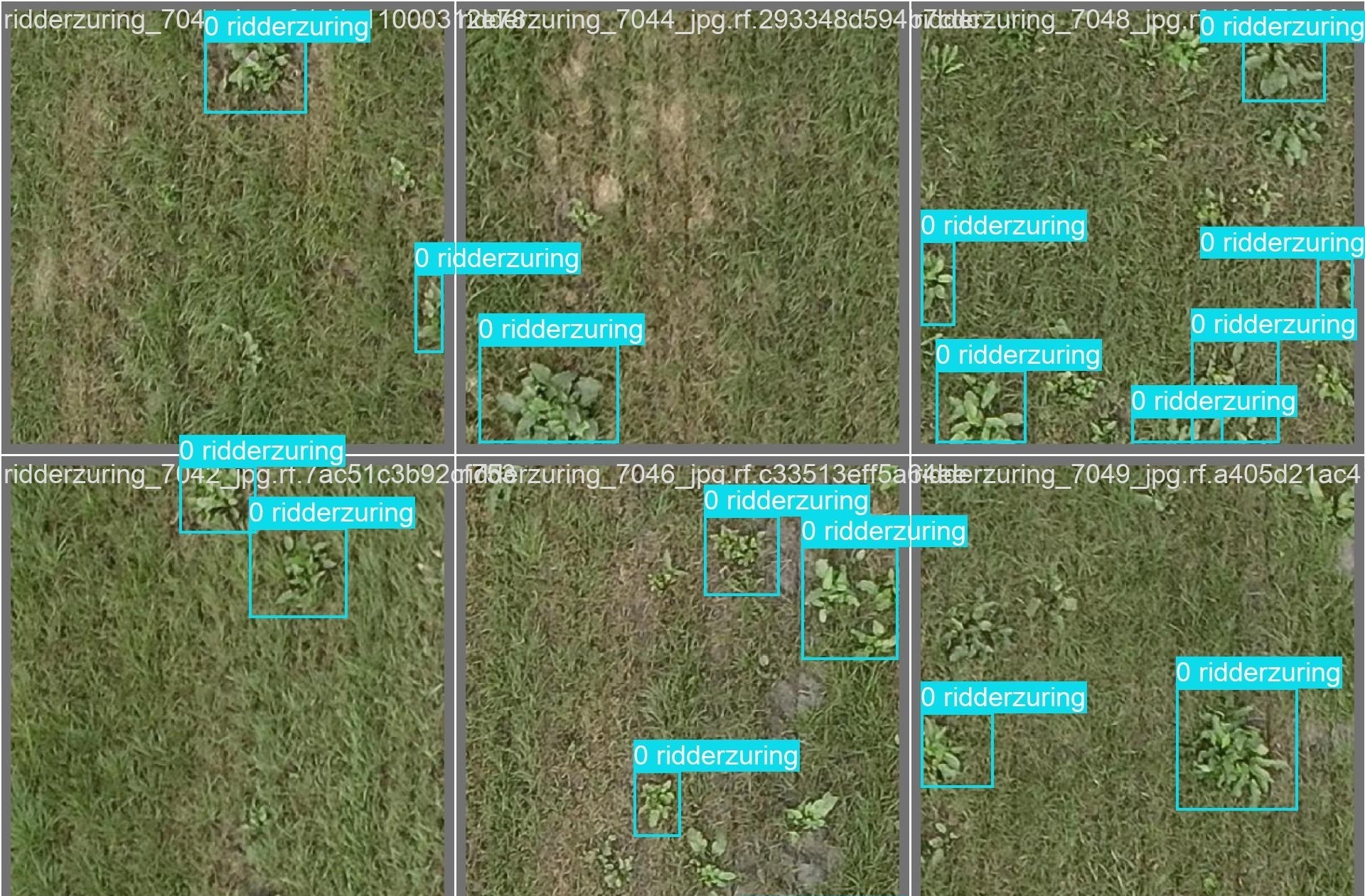}
        \caption{Weed detection and recognition results.}
        \label{result_yolo}
    \end{subfigure}
    \hfill
    \begin{subfigure}{0.55\textwidth}
        \centering
        \includegraphics[width=\linewidth]{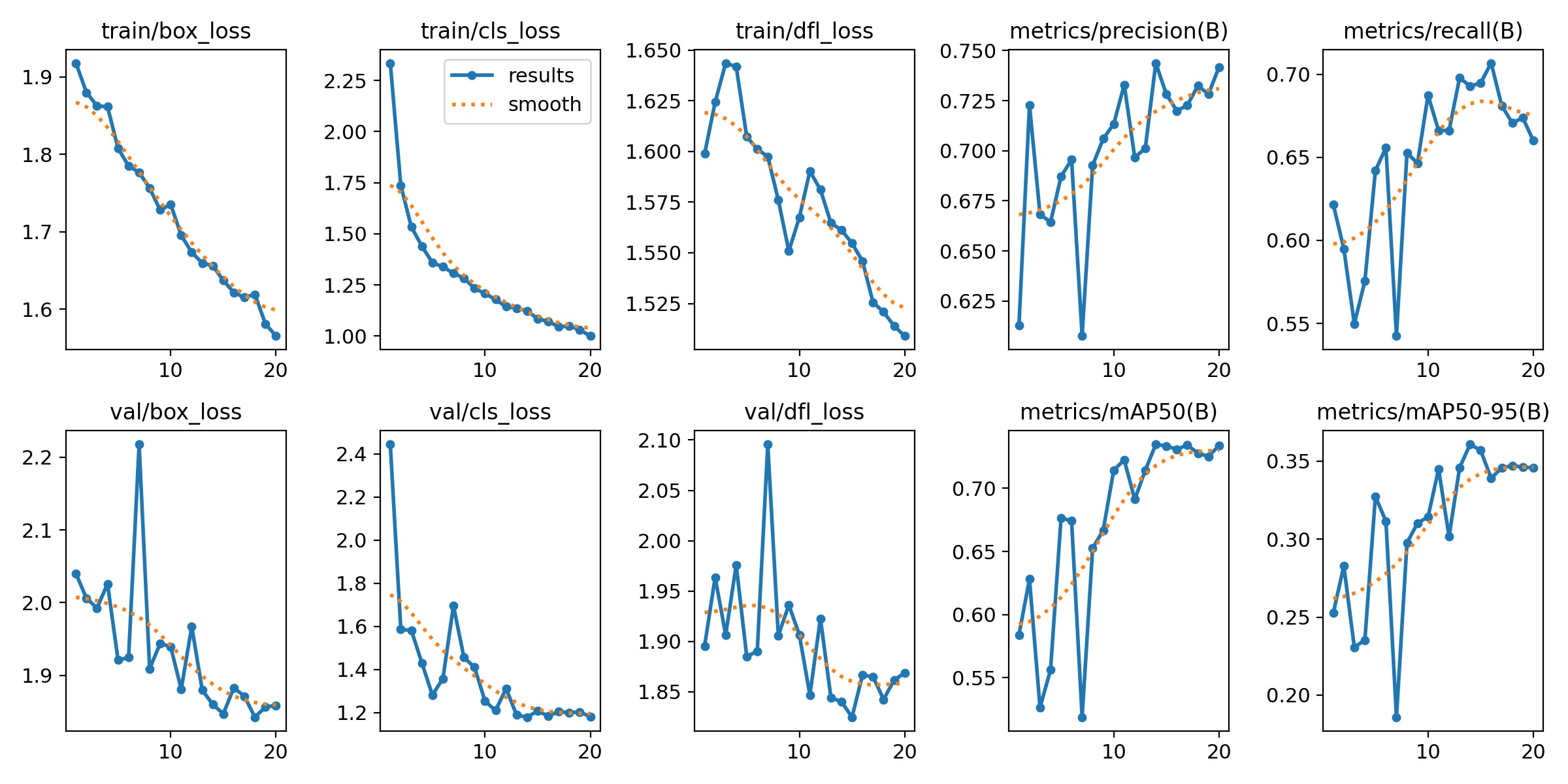}
        \caption{Weed detection and recognition metrics using YOLOv12~\cite{yolov12}.}
        \label{performance_yolo}
    \end{subfigure}
      \hfill
    \begin{subfigure}{0.33\textwidth}
        \centering
        \includegraphics[width=\linewidth]{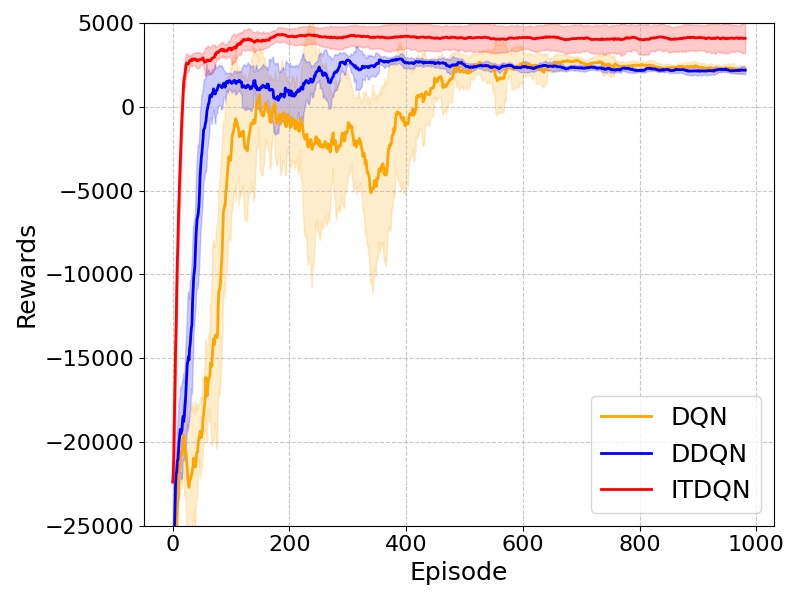}
        \caption{Comparison of reward versus number of episodes using DQN, DDQN, and ITDQN.}
        \label{comparison}
    \end{subfigure}
     \hfill
    \begin{subfigure}{0.32\textwidth}
        \centering
        \includegraphics[width=\linewidth]{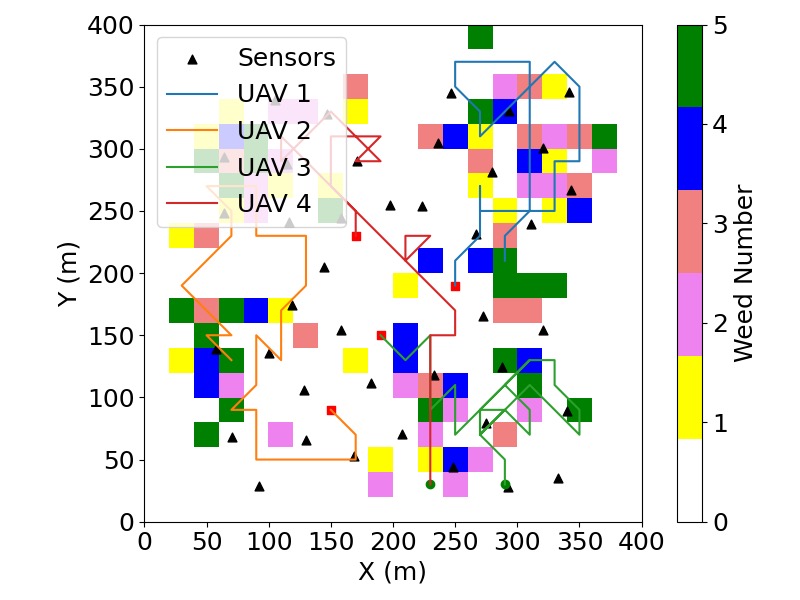}
        \caption{UAV trajectories in synthetic simulation, where the weed density is derived from the weed recognition results shown in (4a).}
        \label{UAV_trajectory_simulation}
    \end{subfigure}
    \hfill
        \begin{subfigure}{0.32\textwidth}
        \centering
        \includegraphics[width=\linewidth]{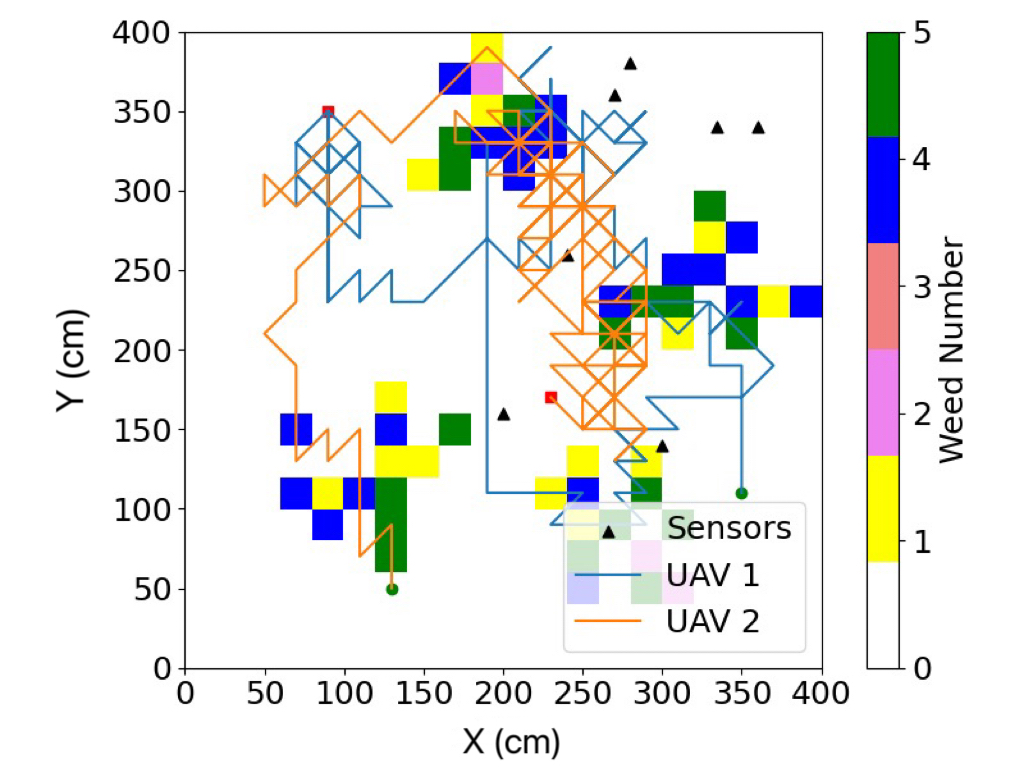}
        \caption{UAV trajectories in real-world demo, where the weed density is represented by markers with different colors shown in (3b).}
        \label{UAV_trajectory_demo}
    \end{subfigure}
    \hfill
        \caption{Experimental Results.}
        \label{fig: experiment_result}
\end{figure*}

In the real-world demonstration, we utilize 2 DJI Tello UAVs flying in a $4~m\times 4~m$ indoor area. 
We use markers with different colors to represent different densities of weeds, and we use 8 Internet of Things (IoT) devices and Raspberry Pis to represent wireless sensors.
Table~\ref{performance_metrics} shows that our proposed ITDQN consistently performs well, with an 85.05\% weed recognition rate and a 79.75\% data collection rate. Furthermore, Fig.~\ref{UAV_trajectory_demo} shows that compared to that in simulation, the UAV trajectories in the real-world demonstration encounter more zig-zag detours, indicating more challenging conditions when UAVs explore in the real-world environments.

\begin{table*}[t]
\caption{Performance metrics in the experiments.}
\label{performance_metrics}
\centering
\setlength{\tabcolsep}{5pt} 
\begin{tabular}{lccccc}
\toprule
\textbf{Algorithm} &
\textbf{Energy (J)} &
\textbf{Recognition (\%)} &
\textbf{Collection (\%)} &
\textbf{Completion (s)} &
\textbf{Inference (ms)} \\
\midrule
\multicolumn{6}{c}{\textbf{Synthetic simulation}}\\
\midrule
ACO   & 4435.17 & 97.50 & 66.75 & 41.00 & 114.71 \\
PSO   & 4867.87 & 97.50 & 83.50 & 45.00 & 130.90 \\
GA    & 5300.57 & 97.50 & 73.75 & 49.00 & 138.09 \\
DQN   & 4220.96 & 75.74 & 91.40 & 34.53 & 5.79  \\
DDQN  & 4669.34 & 75.00 & 91.11 & 38.80 & 6.16  \\
\textbf{ITDQN} & 5608.84 & 79.43 & 98.05 & 47.45 & 6.53  \\
\midrule
\multicolumn{6}{c}{\textbf{Real-world demonstration}}\\
\midrule
\textbf{ITDQN} & 31132.57 & 85.08 & 79.75 & 91.75 & 3.45 \\
\bottomrule
\end{tabular}
\end{table*}

\section{Conclusion}
In this work, we formulate the trajectory planning problem for UAV-based smart farming as an MDP and employ MARL for its solution. In particular, we introduce a novel ITDQN algorithm, which integrates an elite imitation mechanism to lower exploration costs and incorporates a mediator Q-network over a conventional DDQN to enhance performance, training efficiency, and stability. Experimental results conducted in simulated and real-world environments validate the effectiveness of our approach. Furthermore, ITDQN surpasses state-of-the-art baselines, including DQN, DDQN, and heuristics.
In the future, we will evaluate the effectiveness of ITDQN in real-world farmland environments.

\bibliographystyle{unsrt}  
\bibliography{references}

\begin{thebibliography}{10}

\bibitem{big_data_2017}
Sjaak Wolfert, Lan Ge, Cor Verdouw, and Marc-Jeroen Bogaardt.
\newblock Big data in smart farming – a review.
\newblock {\em Agricultural Systems}, 153:69--80, 2017.

\bibitem{europe_2021}
Vasileios Moysiadis, Panagiotis Sarigiannidis, Vasileios Vitsas, and Adel Khelifi.
\newblock Smart farming in europe.
\newblock {\em Computer Science Review}, 39:100345, 2021.

\bibitem{survey_2021}
Godwin Idoje, Tasos Dagiuklas, and Muddesar Iqbal.
\newblock Survey for smart farming technologies: Challenges and issues.
\newblock {\em Computers \& Electrical Engineering}, 92:107104, 2021.

\bibitem{review_2021}
Nahina Islam, Md~Mamunur Rashid, Faezeh Pasandideh, Biplob Ray, Steven Moore, and Rajan Kadel.
\newblock A review of applications and communication technologies for internet of things (iot) and unmanned aerial vehicle (uav) based sustainable smart farming.
\newblock {\em Sustainability}, 13(4), 2021.

\bibitem{farm1}
P.~Parameswari, V.~Sujitha, J.~Surya, B.S.~Aswin Kumar, and S.~Aakash.
\newblock Optimized uav trajectory planning for precision agriculture using wireless sensor networks.
\newblock In {\em 2025 3rd International Conference on Advancements in Electrical, Electronics, Communication, Computing and Automation (ICAECA)}, pages 1--6, 2025.

\bibitem{farm2}
Jahid Choton and William Hsu.
\newblock Coverage path planning in precision agriculture: Algorithms, applications, and key benefits, 12 2024.

\bibitem{farm2.5}
Ravil~I. Mukhamediev, Kirill Yakunin, Margulan Aubakirov, Ilyas Assanov, Yan Kuchin, Adilkhan Symagulov, Vitaly Levashenko, Elena Zaitseva, Dmitry Sokolov, and Yedilkhan Amirgaliyev.
\newblock Coverage path planning optimization of heterogeneous uavs group for precision agriculture.
\newblock {\em IEEE Access}, 11:5789--5803, 2023.

\bibitem{farm3}
Xinyu Liu, Guangquan Li, Haoyuan Yang, Nianru Zhang, Longfei Wang, and Peng Shao.
\newblock Agricultural uav trajectory planning by incorporating multi-mechanism improved grey wolf optimization algorithm.
\newblock {\em Expert Systems with Applications}, 233:120946, 2023.

\bibitem{farm4}
Rick {van Essen}, Eldert {van Henten}, Lammert Kooistra, and Gert Kootstra.
\newblock Adaptive path planning for efficient object search by uavs in agricultural fields.
\newblock {\em Smart Agricultural Technology}, 12:101075, 2025.

\bibitem{farm5}
Haitao Fu, Zheng Li, Weijian Zhang, Yuxuan Feng, Li~Zhu, Xu~Fang, and Jian Li.
\newblock Research on path planning of agricultural uav based on improved deep reinforcement learning.
\newblock {\em Agronomy}, 14(11), 2024.

\bibitem{farm6}
Guoqing Zhang, Jiandong Liu, Wei Luo, Yongxiang Zhao, Ruiyin Tang, Keyu Mei, and Penggang Wang.
\newblock A shortest distance priority uav path planning algorithm for precision agriculture.
\newblock {\em Sensors}, 24(23), 2024.

\bibitem{q-learning}
Christopher J. C.~H. Watkins and Peter Dayan.
\newblock {Q}-learning.
\newblock {\em Machine Learning}, 8(3):279--292, 1992.

\bibitem{analysis_dqn}
J.N. Tsitsiklis and B.~Van~Roy.
\newblock An analysis of temporal-difference learning with function approximation.
\newblock {\em IEEE Transactions on Automatic Control}, 42(5):674--690, 1997.

\bibitem{dqn}
Volodymyr Mnih, Koray Kavukcuoglu, David Silver, Andrei~A. Rusu, Joel Veness, Marc~G. Bellemare, Alex Graves, Martin~A. Riedmiller, Andreas~K. Fidjeland, Georg Ostrovski, Stig Petersen, Charlie Beattie, Amir Sadik, Ioannis Antonoglou, Helen King, Dharshan Kumaran, Daan Wierstra, Shane Legg, and Demis Hassabis.
\newblock Human-level control through deep reinforcement learning.
\newblock {\em Nature}, 518:529--533, 2015.

\bibitem{ddqn}
Hado van Hasselt, Arthur Guez, and David Silver.
\newblock Deep reinforcement learning with double q-learning.
\newblock {\em Proceedings of the AAAI Conference on Artificial Intelligence}, 30(1), Mar. 2016.

\bibitem{duel_dqn}
Ziyu Wang, Tom Schaul, Matteo Hessel, Hado Hasselt, Marc Lanctot, and Nando Freitas.
\newblock Dueling network architectures for deep reinforcement learning.
\newblock In Maria~Florina Balcan and Kilian~Q. Weinberger, editors, {\em Proceedings of The 33rd International Conference on Machine Learning}, volume~48 of {\em Proceedings of Machine Learning Research}, pages 1995--2003, New York, New York, USA, 20--22 Jun 2016. PMLR.

\bibitem{Schaul2015PrioritizedER}
Tom Schaul, John Quan, Ioannis Antonoglou, and David Silver.
\newblock Prioritized experience replay.
\newblock {\em CoRR}, abs/1511.05952, 2015.

\bibitem{noisy_dqn}
Meire Fortunato, Mohammad~Gheshlaghi Azar, Bilal Piot, Jacob Menick, Ian Osband, Alex Graves, Vlad Mnih, R{\'{e}}mi Munos, Demis Hassabis, Olivier Pietquin, Charles Blundell, and Shane Legg.
\newblock Noisy networks for exploration.
\newblock {\em CoRR}, abs/1706.10295, 2017.

\bibitem{c51}
Marc~G. Bellemare, Will Dabney, and R\'{e}mi Munos.
\newblock A distributional perspective on reinforcement learning.
\newblock In {\em Proceedings of the 34th International Conference on Machine Learning - Volume 70}, ICML'17, page 449–458. JMLR.org, 2017.

\bibitem{qr_dqn}
Will Dabney, Mark Rowland, Marc~G. Bellemare, and R\'{e}mi Munos.
\newblock Distributional reinforcement learning with quantile regression.
\newblock In {\em Proceedings of the Thirty-Second AAAI Conference on Artificial Intelligence and Thirtieth Innovative Applications of Artificial Intelligence Conference and Eighth AAAI Symposium on Educational Advances in Artificial Intelligence}, AAAI'18/IAAI'18/EAAI'18. AAAI Press, 2018.

\bibitem{idqn}
Will Dabney, Georg Ostrovski, David Silver, and Remi Munos.
\newblock Implicit quantile networks for distributional reinforcement learning.
\newblock In Jennifer Dy and Andreas Krause, editors, {\em Proceedings of the 35th International Conference on Machine Learning}, volume~80 of {\em Proceedings of Machine Learning Research}, pages 1096--1105. PMLR, 10--15 Jul 2018.

\bibitem{rainbow_dqn}
Matteo Hessel, Joseph Modayil, Hado van Hasselt, Tom Schaul, Georg Ostrovski, Will Dabney, Dan Horgan, Bilal Piot, Mohammad Azar, and David Silver.
\newblock Rainbow: combining improvements in deep reinforcement learning.
\newblock In {\em Proceedings of the Thirty-Second AAAI Conference on Artificial Intelligence and Thirtieth Innovative Applications of Artificial Intelligence Conference and Eighth AAAI Symposium on Educational Advances in Artificial Intelligence}, AAAI'18/IAAI'18/EAAI'18. AAAI Press, 2018.

\bibitem{es_dqn}
Adrian Ly, Richard Dazeley, Peter Vamplew, Francisco Cruz, and Sunil Aryal.
\newblock Elastic step dqn: A novel multi-step algorithm to alleviate overestimation in deep qnetworks, 2022.

\bibitem{dr_dqn}
Matthew~J. Hausknecht and Peter Stone.
\newblock Deep recurrent q-learning for partially observable mdps.
\newblock {\em CoRR}, abs/1507.06527, 2015.

\bibitem{weed_2017}
Philipp Lottes, Raghav Khanna, Johannes Pfeifer, Roland Siegwart, and Cyrill Stachniss.
\newblock Uav-based crop and weed classification for smart farming.
\newblock In {\em 2017 IEEE International Conference on Robotics and Automation (ICRA)}, pages 3024--3031, 2017.

\bibitem{weed_dataset}
Jai Dalmotra.
\newblock Weed detection, 2023.

\bibitem{yolov12}
Yunjie Tian, Qixiang Ye, and David Doermann.
\newblock Yolov12: Attention-centric real-time object detectors, 2025.

\bibitem{aco}
Marco Dorigo, Mauro Birattari, and Thomas Stutzle.
\newblock Ant colony optimization.
\newblock {\em IEEE Computational Intelligence Magazine}, 1(4):28--39, 2006.

\bibitem{pso}
J.~Kennedy and R.~Eberhart.
\newblock Particle swarm optimization.
\newblock In {\em Proceedings of ICNN'95 - International Conference on Neural Networks}, volume~4, pages 1942--1948 vol.4, 1995.

\bibitem{ga}
Seyedali Mirjalili.
\newblock Genetic algorithm.
\newblock {\em Handbook of Machine Learning}, 2018.

\end{thebibliography}

\end{document}